# HYBRID DATA CLUSTERING APPROACH USING K-MEANS AND FLOWER POLLINATION ALGORITHM

R.Jensi[1] and G.Wiselin Jiji[2]

[1]Department of CSE, Manomanium Sundaranar University, India
[2]Dr.Sivanthi Aditanar College of Engineering, India

## ABSTRACT

*Data clustering is a technique for clustering set of objects into known number of groups. Several approaches are widely applied to data clustering so that objects within the clusters are similar and objects in different clusters are far away from each other. K-Means, is one of the familiar center based clustering algorithms since implementation is very easy and fast convergence. However, K-Means algorithm suffers from initialization, hence trapped in local optima. Flower Pollination Algorithm (FPA) is the global optimization technique, which avoids trapping in local optimum solution. In this paper, a novel hybrid data clustering approach using Flower Pollination Algorithm and K-Means (FPAKM) is proposed. The proposed algorithm results are compared with K-Means and FPA on eight datasets. From the experimental results, FPAKM is better than FPA and K-Means.*

## 1. INTRODUCTION

Data clustering [4] [6] is an unsupervised learning technique in which class labels are not known in advance. The purpose of clustering is to partition a set of objects into clusters or groups so that the objects within the cluster are more similar to each other, while objects in different clusters are far away from each other. In past decades, many nature-inspired evolutionary algorithms have been developed for solving most engineering design optimization problems, which are highly nonlinear, involving many design variables and complex constraints. These metaheuristic algorithms are attracted very much because of the global search capability and take less time to solve real world problems. Nature-inspired algorithms [2] [3] imitate the behaviours of the living things in the nature, so they are also called as Swarm Intelligence (SI) algorithms.

Evolutionary algorithms (EAs) were the initial stage of such optimization methods [35]. Genetic Algorithm (GA) [6] and Simulated Annealing (SA) [7] are popular examples for EAs. In the early 1970s, Genetic algorithm was developed by John Holland, which inspired by biological evolution such as reproduction, mutation, crossover and selection. Simulated annealing (SA) was developed from inspiration by annealing in metallurgy, a technique involving heating and cooling of a material to increase the size of its crystals and reduce their defects.

The rising body of Swarm Intelligence(SI) [2] [3] metaheuristic algorithms include Particle Swarm Optimization (PSO) [1] [5], Ant Colony Optimization (ACO) [14], Glowworm Swarm Optimization (GSO) [8], Bacterial Foraging Optimization (BFO) [9-10], the Bees Algorithm [31], Artificial Bee Colony algorithm (ABC) [25][28-29], Biogeography-based optimization (BBO) [30], Cuckoo Search (CS) [26-27], Firefly Algorithm (FA) [32-33], Bat Algorithm (BA) [20] and flower pollination algorithm[19].

15

Swarm Intelligence system holds a population of solutions, which are changed through random selection and alterations of these solutions. The way, the system differs depends on the generation of new solutions, random selection procedure and candidate solution encoding technique. Particle Swarm Optimization (PSO) was developed in 1995 by Kennedy and Eberhart simulating the social behaviour of bird flock or fish school. Ant Colony Optimization, introduced by Dorigo, imitates the food searching paths of ants in nature. Glowworm Swarm Optimization (GSO) was introduced by Krishnanand and Ghose in 2005 based on the behaviour of glow worms. Bacterial foraging optimization algorithm was developed based on the foraging behaviour of bacteria such as E.coli and M.xanthus. The Bees Algorithm was developed by Pham DT in 2005 imitating the food foraging behaviour of honey bee colonies. Artificial bee colony algorithm was developed by Karaboga, being motivated from food foraging behaviour of bee colonies. Biogeography-based optimization (BBO) was introduced in 2008 by Dan Simon inspired by biogeography, which is the study of the distribution of biological species through space and time. Cuckoo search was developed by Xin-she Yang and Subash Deb in 2009 being motivated by the brood parasitism of cuckoo species by laying their eggs in the nests of other host birds. Firefly algorithm was introduced by Xin-She Yang inspired by the flashing behaviour of fireflies. The primary principle for a firefly's flash is to act as an indicator system to draw other fireflies. Bat algorithm was developed in 2010 by Xin-She Yang based on the echolocation behaviour of microbats. Flower pollination algorithm was developed by Xin-She Yang in 2012 motivated by the pollination process of flowering plants.

The remainder of this paper is organized as follows. Section 2 presents some of the previous proposed research work on data clustering. K-Means algorithm and Flower Pollination algorithm is presented in Section 3 and Section 4 respectively. Then in Section 5 proposed algorithm is explained. Section 6 discusses experimental results and Section 7 concludes the paper with fewer discussions.

## 2. RELATED WORK

Van, D.M. and A.P. Engelbrecht. (2003) [5] proposed data clustering approach using particle swarm optimization. The author proposed two approaches for data clustering. The first approach is, PSO, in which the optimal centroids are found and then these optimal centroids were used as a seed in K-means algorithm and the second approach is, the PSO was used to refine the clusters formed by K-means. The two approaches were tested and the results show that both PSO clustering techniques have much potential.

Ant Colony Optimization (ACO) method for clustering is presented by Shelokar et al. (2004) [14]. In [14], the authors employed distributed agents that imitate the way real-life ants find the shortest path from their nest to a food source and back. The results obtained by ACO can be considered viable and is an efficient heuristic to find near-optimal cluster representation for the clustering problem.

Kao et al. (2008) [22] proposed a hybridized approach that combines PSO technique, Nelder–Mead simplex search and the K-means algorithm. The performance of K-NM-PSO is compared with PSO, NM-PSO, K-PSO and K-means clustering and it is proved that K-NM-PSO is both strong and suitable for handling data clustering.

Maulik and Mukhopadhyay (2010) [7] also presented a simulated annealing approach to clustering. They combined their heuristic with artificial neural networks to improve solution quality and the similarity criteria, which used DB cluster validity index. Karaboga and Ozturk (2011) [15] presented a new clustering approach using Artificial Bee Colony (ABC) algorithm which simulates the food foraging behaviour of a honey bee swarm. The performance is compared with PSO and other classification techniques. The simulation results show that the ABC algorithm is superior to other algorithms.

Zhang et al. (2010) [23] presented the artificial bee colony (ABC) as a state-of-the-art approach to clustering. Deb's rules are used to tackle infeasible solutions instead of the greedy selection process usually used in the ABC algorithm. When they tested their algorithm, they found very encouraging results in terms of effectiveness and efficiency.

In [16] (2012), X. Yan et al presented a new data clustering algorithm using hybrid artificial bee colony (HABC). The genetic algorithm crossover operator was introduced to ABC to enhance the information exchange between bees. The HABC algorithm achieved better results.

Tunchan Cura. (2012) [19] presented a new PSO approach to the data clustering and the algorithm was tested using two synthetic datasets and five real datasets. The results show that the algorithm can be applied to clustering problem with known and unknown number of clusters. Senthilnath, J., Omkar, S.N. and Mani, V. (2011) [13] presented data clustering using firefly algorithm. They measured the performance of FA with respect to supervised clustering problem and the results show that algorithm is robust and efficient.

M.Wan and his co-authors (2012) [17] presented data clustering using Bacterial Foraging Optimization (BFO). The algorithm proposed by these researchers was tested on several well-known benchmark data sets and Compared three clustering technique. The author concludes that the algorithm is effective and can be used to handle data sets with various cluster sizes, densities and multiple dimensions.

J. Senthilnatha, Vipul Dasb, Omkara, V. Mani, (2012) [18] proposed a new data clustering approach using Cuckoo search with levy flight. Levy flight is heavy-tailed which ensures that it covers output domain efficiently. The author concluded that the proposed algorithm is better than GA and PSO.

## 3. K-MEANS ALGORITHM

K-Means Clustering algorithm is fast and easy to implement. Due to its simplicity, K-Means clustering is heavily used. The process of clustering using K-Means is as follows:

Let $O = \{o_1, o_2, \ldots, o_n\}$ be a set of n data objects to be partitioned and each data object $o_i$, i=1,2,…,n is represented as $o_i = \{o_{i1}, o_{i2}, \ldots, o_{im}\}$ where $o_{im}$ represents $m^{th}$ dimension value of data object i.

The output clustering algorithm is a set of K partitions $P = \{P_1, P_2, \ldots, P_k \mid \forall k : P_k \neq \emptyset$ and $\forall l \neq k : P_k \cap P_l = \emptyset\}$ such that objects within the clusters are more similar and dissimilar to objects in different clusters. These similarities are measured by some optimization criterion, especially total within-cluster variance or the total mean-square quantization error (MSE) which is defined as:

$$\text{Min} \sum_{j=1}^{K} \sum_{i=1}^{n} w_{ij} E(o_i, p_j) \tag{1}$$

where $p_j$ represents a $j^{th}$ cluster center, E is the distance measure between a data object $o_i$ and a cluster center $p_j$, $w_{ij} \in \{0,1\}$ denotes that object i belongs to cluster j if $w_{ij}=1$ (otherwise $w_{ij}=0$). In this paper Euclidean distance is used as distance metric which is defined as follows:

$$E(o_i, p_j) = \sqrt{\sum_{m=1}^{M}(o_{im} - p_{jm})^2} \quad (2)$$

where,
$p_j$ is cluster center for a cluster j and is calculated as follows:

$$p_j = \frac{1}{n_j} \sum_{o_i \in p_j} o_i \quad (3)$$

where, $n_j$ is the total number of objects in cluster j.
The K-Means algorithm is defined in fig. (1).

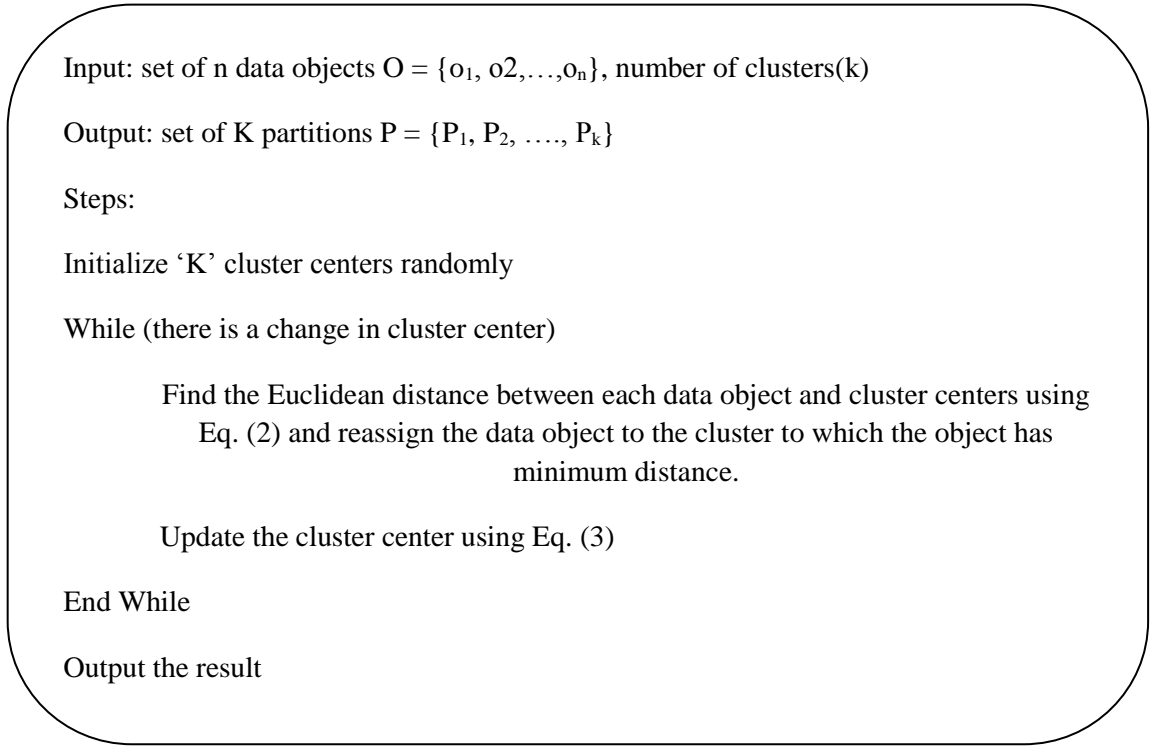

Input: set of n data objects O = {o1, o2,…,on}, number of clusters(k)

Output: set of K partitions P = {P1, P2, …., Pk}

Steps:

Initialize 'K' cluster centers randomly

While (there is a change in cluster center)

    Find the Euclidean distance between each data object and cluster centers using Eq. (2) and reassign the data object to the cluster to which the object has minimum distance.

    Update the cluster center using Eq. (3)

End While

Output the result

Figure 1. K-Means Algorithm

## 4. FLOWER POLLINATION ALGORITHM (FPA)

Flower Pollination Algorithm (FPA) is a global optimization algorithm, which was introduced by Xin-She Yang in 2012 [19], inspired by the pollination process of flowers. There are two key steps in FPA. One is global pollination and the other is local pollination. In the global pollination step, insects fly and move in a longer distance and the fittest is represented by g*. The flower pollination process with longer move distance is carried out with levy flights. Mathematically, the global pollination process is represented as

$$x_i^{t+1} = x_i^t + L(x_i^t - g^*) \quad (4)$$

where,
$x_i^t$     - solution vector at iteration t
$x_i^{t+1}$     - solution vector at iteration t+1
$g^*$     - best solution
L     - step size.

The step size L is drawn from Levy flight distribution [35],

$$L \sim \frac{\lambda \Gamma(\lambda) \sin\left(\frac{\pi\lambda}{2}\right)}{\pi} \frac{1}{s^{1+\lambda}}, \qquad (s \gg s0 > 0). \qquad (5)$$

where,

$\Gamma(\lambda)$- Standard gamma function and $\lambda = 3/2$.

In the local pollination step, self-pollination is depicted. It is mathematically represented by

$$x_i^{t+1} = x_i^t + \epsilon \, (x_j^t - x_k^t) \qquad (6)$$

where,
$x_i^t$     - solution vector at iteration t
$x_i^{t+1}$ - solution vector at iteration t+1
$\epsilon$  - random uniformly distributed number between [0,1]
$j, k$      - randomly selected indices.

To perform global and local pollination process, a switch probability is used to switch between global and local scale. The FPA is summarized in Fig.(2).

```
Initialize a population of n flowers with randomly generated solutions
Evaluate the solutions
Find the current best solution among all solutions in the initial population
Assign a switch probability p ε [0, 1]
t=0
Define maximum number iteration (maxIter)
While (t<maxIter)
        For each flower/solution in the population
                If rand <p
                        Perform global pollination using Eq. (4)
                Else
                        Perform local pollination using Eq. (6)
                End if
                Evaluate new solutions
                Update the solution if new is better than current
        End for
        Find the current fittest solution g* and update g*
        t=t+1
End while
Output the best solution
```

Figure 2. Flower Pollination Algorithm

## 5. FLOWER POLLINATION ALGORITHM WITH K-MEANS (FPAKM)

In this paper, the flower pollination algorithm is integrated with K-Means (FPAKM) to form a hybrid clustering algorithm, which gives all functionalities of FPA and K-Means. If the current best solution does not improve in a predetermined number of trials, a local search around current best solution is made. The maximum of trial is the limit value. The objective function used is total mean-square quantization error (MSE). The proposed algorithm is given in fig.(3).

```
Initialize a population of n flowers with randomly generated
solutions
Evaluate the solution using Eq. (1)
Find current best solution among all solutions in the initial
population
Assign a switch probability p ε [0, 1]
t=0, trial=0
Define maximum number iteration (maxIter)
While (t<maxIter)
        For each flower/solution in the population
          If trial<limit
              If rand <p
                    Perform global pollination using Eq.
(4)
              Else
                    Perform local pollination using Eq. (6)
              End if
              Evaluate new solutions using Eq. (1)
              Update the solution if new is better than current
          Else [local search]
              Take the current solution as initial seed of K-
              Means clustering algorithm.
               Recalculate the cluster center and update
solution.
          End if
        End for
          Find the current fittest solution g*
          If no change in g*
              trial=trial+1
          Else
                update g*
                trial=0.
          End if
          t=t+1
End while
Output the best solution
```

Figure 3. Flower Pollination Algorithm with K-Means

## 6. EXPERIMENTAL RESULTS AND DISCUSSION

The K-Means, Flower Pollination Algorithm (FPA) and proposed algorithm (FPAKM) are written in Matlab 8.3 and executed in a Windows 7 Professional OS environment using Intel i3, 2.30 GHz, 2 GB RAM. Flower Pollination Algorithm (FPA) matlab code is available in [34]. For comparison easier, FPA, FPAKM are executed with the parameters except limit value as shown in Table 1.

Table 1. FPA, FPAKM Control Parameters and its values

| Parameter | Value |
|---|---|
| Max Generation | 2000 |
| Number of flowers(N) | 20 |
| Limit | 2 |
| Switch probability (p) | 0.8 |

### 6.1 Dataset Description

To evaluate the performance of proposed algorithm, eight datasets have been used. One is artificial dataset drawn from Kao et al. (2008). The remaining seven datasets, namely, iris, thyroid, wine, Contraceptive Method Choice (CMC), crude oil and glass, are collected from [36]. The artificial dataset contains samples drawn from five independent uniform distributions with ranges of (85,100), (70, 85), (55, 70), (40, 55) and (25, 40). The eight datasets used in this paper is described in Table 2.

Table 2 Dataset Characteristics

| S.No | Dataset Name | # of attributes | # of classes | # of instances |
|---|---|---|---|---|
| 1 | Artset1 | 3 | 5 | 250 |
| 2 | Iris | 4 | 3 | 150 |
| 3 | Wine | 13 | 3 | 178 |
| 4 | Glass | 9 | 6 | 214 |
| 5 | Cancer | 9 | 2 | 683 |
| 6 | Thyroid | 5 | 3 | 215 |
| 7 | Contraceptive Method Choice (CMC) | 9 | 3 | 1473 |
| 8 | Crude Oil | 5 | 3 | 56 |

### 6.2 Performance Evaluation

The quality of clustering algorithms is measured using objective function value and F-measure. The smaller the objective function value is, the quality of clustering will be higher.

The F-measure employs the ideas of precision and recall values used in information retrieval. The *precision P(i,j)* and *recall R(i,j)* of each cluster j for each class i are calculated as

$$P(i,j) = \frac{\beta_{ij}}{\beta_j} \qquad (7)$$

$$R(i,j) = \frac{\beta_{ij}}{\beta_i} \qquad (8)$$

where,

$\beta_i$ : is the number of members of class i
$\beta_j$ : is the number of members of cluster j
$\beta_{ij}$: is the number of members of class i in cluster j

The corresponding *F-measure F(i,j)* is given in Eq. (9):

$$F(i,j) = \frac{2 * P(i,j) * R(i,j)}{P(i,j) + R(i,j)} \qquad (9)$$

Then the *F-measure* of a class i can be defined as

$$\text{Ftot} = \sum_i \frac{\beta_i}{n} \max_j (F(i,j)) \qquad (10)$$

where, n is the total number of data objects in the collection. In general, the larger the F-measure gives the better clustering result.

## 6.3 Results Discussion

In this paper, to compare the performance of proposed algorithm, each algorithm has been run for 10 times and the best, worst, average and the standard deviation of each algorithms' objective function values and F-measure values are given in table 3. The best values are shown in bold face.

For Artificial dataset Art1, the FPAKM algorithm obtains the best objective function value and F-measure value is 1 for FPA and FPAKM algorithm, while K-Means algorithm also gives maximum F-measure. For all datasets except cmc data, the FPAKM outperforms other two algorithms and it results the best value in terms of objective function value and F-measure. For datasets Art1, Iris, Wine, cancer, thyroid, crude oil, the FPA obtains maximum F-measure value over 10 runs. But FPAKM always obtains the best results compared to FPA and K-Means.

Table 3. Objective function values and F-measure values of three clustering algorithms

| Data set | Algorithm | Objective Function Value | | | | F-measure | | | |
|---|---|---|---|---|---|---|---|---|---|
| | | Best | Worst | Average | Std | $F_{min}$ | $F_{max}$ | $F_{avg}$ | $F_{std}$ |
| Art1 | K-Means | 1750.355 | 2379.748 | 1813.295 | 199.032 | 0.789 | 1.000 | 0.979 | 0.067 |
| | FPA | 1747.951 | 1749.414 | 1748.732 | 0.487 | 1.000 | 1.000 | 1.000 | 0.000 |
| | FPAKM | 1747.725 | 1748.175 | 1747.943 | 0.162 | 1.000 | 1.000 | 1.000 | 0.000 |
| Iris | K-Means | 97.326 | 123.850 | 102.48 | 10.854 | 0.659 | 0.892 | 0.846 | 0.091 |
| | FPA | 96.656 | 96.748 | 96.693 | 0.034 | 0.892 | 0.899 | 0.898 | 0.002 |
| | FPAKM | 96.664 | 96.682 | 96.673 | 0.005 | 0.899 | 0.899 | 0.899 | 0.000 |
| Wine | K-Means | 16555.679 | 18436.952 | 16931.934 | 793.214 | 0.636 | 0.715 | 0.699 | 0.033 |
| | FPA | 16296.401 | 16310.213 | 16301.697 | 3.918 | 0.719 | 0.729 | 0.722 | 0.004 |
| | FPAKM | 16292.865 | 16294.951 | 16293.623 | 0.636 | 0.729 | 0.729 | 0.729 | 0.000 |
| Glass | K-Means | 215.915 | 253.719 | 226.071 | 14.441 | 0.468 | 0.561 | 0.537 | 0.031 |
| | FPA | 220.499 | 228.698 | 223.423 | 2.779 | 0.487 | 0.542 | 0.518 | 0.017 |
| | FPAKM | 211.482 | 214.354 | 211.923 | 0.866 | 0.550 | 0.561 | 0.558 | 0.004 |
| Cancer | K-Means | 2986.961 | 2988.428 | 2987.988 | 0.708 | 0.960 | 0.962 | 0.961 | 0.001 |
| | FPA | 2965.041 | 2971.477 | 2966.848 | 2.091 | 0.963 | 0.965 | 0.964 | 0.001 |
| | FPAKM | 2964.648 | 2965.445 | 2964.994 | 0.235 | 0.965 | 0.965 | 0.965 | 0.000 |
| Thyroid | K-Means | 1984.705 | 2017.046 | 1992.560 | 10.245 | 0.652 | 0.860 | 0.797 | 0.073 |
| | FPA | 1866.678 | 1881.438 | 1870.001 | 4.207 | 0.619 | 0.670 | 0.636 | 0.018 |
| | FPAKM | 1867.862 | 1870.684 | 1868.967 | 0.926 | 0.619 | 0.665 | 0.636 | 0.014 |
| cmc | K-Means | 5542.182 | 5545.333 | 5543.701 | 1.604 | 0.402 | 0.406 | 0.404 | 0.002 |
| | FPA | 5541.565 | 5571.853 | 5555.147 | 10.075 | 0.400 | 0.407 | 0.402 | 0.002 |
| | FPAKM | 5534.763 | 5536.020 | 5535.529 | 0.363 | 0.401 | 0.401 | 0.401 | 0.000 |
| Crude Oil | K-Means | 279.485 | 279.743 | 279.666 | 0.125 | 0.674 | 0.658 | 0.663 | 0.008 |
| | FPA | 277.258 | 277.885 | 277.535 | 0.211 | 0.674 | 0.705 | 0.699 | 0.011 |
| | FPAKM | 277.251 | 277.313 | 277.281 | 0.019 | 0.705 | 0.705 | 0.705 | 0.000 |

# 7. CONCLUSION

This paper presents a hybrid data clustering algorithm (FPAKM) based on the K-Means and Flower Pollination algorithm. The results obtained by the proposed algorithm are compared with K-Means and flower pollination algorithm. It is revealed that the proposed algorithm finds optimal cluster centers, hence the F-measure value is increased. In mere future, this algorithm can be applied to solve other optimization problems.